# Translating between Horn Representations
# and their Characteristic Models

**Roni Khardon**                                                    RONI@DAS.HARVARD.EDU

*Aiken Computation Lab., Harvard University*
*Cambridge, MA 02138 USA*

## Abstract

Characteristic models are an alternative, model based, representation for Horn expressions. It has been shown that these two representations are incomparable and each has its advantages over the other. It is therefore natural to ask what is the cost of translating, back and forth, between these representations. Interestingly, the same translation questions arise in database theory, where it has applications to the design of relational databases. This paper studies the computational complexity of these problems.

Our main result is that the two translation problems are equivalent under polynomial reductions, and that they are equivalent to the corresponding decision problem. Namely, translating is equivalent to deciding whether a given set of models is the set of characteristic models for a given Horn expression.

We also relate these problems to the hypergraph transversal problem, a well known problem which is related to other applications in AI and for which no polynomial time algorithm is known. It is shown that in general our translation problems are at least as hard as the hypergraph transversal problem, and in a special case they are equivalent to it.

## 1. Introduction

The traditional form of representing knowledge in AI is through logical formulas (McCarthy, 1958; McCarthy & Hayes, 1969), where all the logical conclusions of a given formula are assumed to be accessible to an agent. Recently, an alternative way of capturing such information has been developed (Kautz, Kearns, & Selman, 1995; Khardon & Roth, 1994). Instead of using a logical formula, the knowledge representation is composed of a particular subset of its models, the set of *characteristic models*. This set retains all the information about the formula, and is useful for various reasoning tasks. In particular, using model evaluation with the set of characteristic models, one can deduce whether another formula, a query presented to an agent, is implied by the knowledge or not. While characteristic models exist for arbitrary propositional formulas, in this paper we limit our attention to logical formulas which are in Horn form and to their representation as characteristic models.

The characteristic models of Horn formulas have been shown to be useful. There is a linear time deduction algorithm using this set, and abduction can be performed in polynomial time, while using formulas it is NP-Hard (Kautz et al., 1995). Furthermore, an algorithm for default reasoning using characteristic models has been developed, for cases where formula based algorithms are not known (Khardon & Roth, 1995). Hence, the question arises, whether one can efficiently translate a Horn formula into its set of characteristic models and then use this set for the reasoning task. We denote this translation problem by CCM (for Computing Characteristic Models).

On the other hand, given a set of assignments, it might be desirable to find the underlying structure behind this set of models. This is the case when one is trying to learn the structure of the world using a set of examples. This problem has been studied before under the name Structure Identification (Dechter & Pearl, 1992; Kautz et al., 1995; Kavvadias, Papadimitriou, & Sideri, 1993). Technically, the problem seeks an efficient translation from a set of characteristic models into a Horn expression that explains it. We denote this translation problem by SID (for Structure Identification).

Interestingly, the same constructs appear in the theory of relational databases. As shown in a companion paper (Khardon, Mannila, & Roth, 1995), there is a correspondence between Horn expressions and Functional Dependencies, and a correspondence between characteristic models and an Armstrong relation. The equivalent question of translating between functional dependencies and Armstrong relations has been studied before (Beeri, Dowd, Fagin, & Statman, 1984; Mannila & Raiha, 1986; Eiter & Gottlob, 1991; Gottlob & Libkin, 1990) and is relevant for the design of relational databases (Mannila & Raiha, 1986). While this paper does not discuss the problems in the database domain, some of the results presented here can be alternatively derived from previous results in database theory using the above mentioned equivalence. (We identify those precisely, later on.) However, this paper makes these results more accessible without resorting to any results in database theory, and with simpler proofs. On the other hand some new results are presented, which resolve a question which was open both in AI and in the database domain.

## 1.1 An Example

Let us introduce the problems in question through an example. Suppose the world has 4 attributes denoted $a, b, c, d$, each taking a value in $\{0, 1\}$ to denote whether it is "on" or "off", and our knowledge is given by the following constraints:

$$W = (bc \rightarrow d)(cd \rightarrow b)(bc \rightarrow a).$$

Then $W$ is a Horn expression and it is normally used to decide whether certain constrains are implied by it or not. For example $W \models (cd \rightarrow a)$, and $W \not\models (bd \rightarrow a)$, where the symbol $\models$ stands for implication. This is normally performed by deriving a proof for the constraint in question. If no such proof exists then implication does not hold. In our example we would notice that $(cd \rightarrow b)$, and therefore $(cd \rightarrow bc \rightarrow a)$. As for $(bd \rightarrow a)$, we would fail to find a proof and therefore conclude that it is not implied by $W$. This general approach is called theorem proving, and is efficient for Horn expressions (Dowling & Gallier, 1984).

An alternative approach is to check the implication relation by model checking. Implication is defined as follows: $W \models \alpha$ if every model of $W$ is also a model of $\alpha$ (where $x \in \{0, 1\}^n$ is a model of an expression $f$ if $f$ is evaluated to "truth" on $x$). So to decide whether $W \models \alpha$ we can simply use all the models of $W$, and check, one by one, whether any of them does not satisfy $\alpha$. In our example $W$ has 11 models:

$$models(W) = \{0000, 0001, 0010, 0100, 0101, 1000, 1001, 1010, 1100, 1101, 1111\}$$

(where the assignments denote the values assigned to $abcd$ correspondingly), and we would have to test $\alpha$ on every one of them. Unfortunately, in general the number of models may be very large, exponential in the number of variables, and therefore this procedure will not be efficient.

The question arises therefore, whether there is a small subset of models which still guarantees correct results when used with the model checking procedure. Such a subset is called the set of *characteristic models* of $W$ and its existence has been proved (Kautz et al., 1995; Khardon & Roth, 1994). In our example this set is:

$$char(W) = \{0010, 0101, 1001, 1010, 1100, 1101, 1111\},$$

so it includes 7 out of the 11 models of $W$. Model checking with this set is guaranteed to produce correct results for any $\alpha$ which is a Horn expression, and using a slightly more complicated algorithm one can answer correctly for every $\alpha$ (Kautz et al., 1995). In our example, it is easy to check that $(cd \rightarrow a)$ is evaluated to "truth" on all the assignments in $char(W)$ and that $(bd \rightarrow a)$ is falsified by 0101.

The utility of these representations, Horn expressions and characteristic models, is not comparable. Each of these representations has its advantages over the other. First, the size of these representations is incomparable. There are short Horn expressions for which the set of characteristic models is of exponential size, and vice versa, there are also exponential size Horn expressions for which the set of characteristic models is small (Kautz et al., 1995). The representations also differ in the services which they support. On one hand, Horn expressions are more comprehensible. On the other hand characteristic models are advantageous in that they allow for efficient algorithms for abduction and default reasoning. In this paper we are asking how hard it is to translate between these representations, so as to enjoy the benefits of both.

## 1.2 Overview of the Paper

In this paper we study the complexity of the translation problems CCM and SID. For these problems, the output may be exponentially larger than the input. Therefore, it is appropriate to ask whether there are algorithms which can perform the above tasks in time which is polynomial in both the input size and the output size. These are called output polynomial algorithms.

Before starting our investigation we note that it has been shown (Kautz et al., 1995) that using the set of characteristic models one can answer abduction queries related to $H$ in polynomial time, while given the formula $H$ it is NP-Hard to perform abduction (Selman & Levesque, 1990). This however does not imply that computing the set of characteristic models is NP-Hard since the construction in the proof yields a Horn formula whose set of characteristic models is of exponential size.

Our main result says that CCM and SID are equivalent to each other, and are also equivalent to the corresponding decision problem. The problem of Characteristic Models Identification (CMI), is the problem of deciding, given a Horn expression $H$ and a set of models $G$, whether $G = char(H)$. We show that CCM, SID, and CMI are equivalent under polynomial reductions. Namely, the translation problems are solvable in polynomial time if and only if the decision problem is solvable in polynomial time. These are new results which have immediate corollaries in the database domain.

We then show a close relationship between these problems and the Hypergraph Transversal Problem (HTR). Given a hypergraph $G$ a transversal of its edges is a set of nodes which touches every edge in the graph. In the HTR problem one is given a hypergraph as an input, and is required to compute the set of minimal transversals of its edges.

The HTR problem has a lot of equivalent manifestations which appear in various branches of computer science. Examples in AI include computing abductive diagnoses (Reiter, 1987), enumerating prime implicants in ATMS (Reiter & De Kleer, 1987), and Horn approximations (Kavvadias et al., 1993) which are closely related to characteristic models. Other areas include database theory (Mannila & Raiha, 1986), Boolean complexity, and distributed systems (Eiter & Gottlob, 1991). A comprehensive study of these problems is presented by Eiter and Gottlob (1994). HTR is also equivalent to the problem of dualization of monotone Boolean expressions, which is the form in which we present it here. This problem, requires translation between the CNF and DNF representations of monotone functions.

The complexity of the HTR problem has been studied before (Fredman & Khachiyan, 1994; Eiter & Gottlob, 1994; Kavvadias et al., 1993) and is still an open question. On one hand a class of problems which are "HTR complete" has been defined and studied (Eiter & Gottlob, 1994). This class includes many problems from various application areas which are equivalent to HTR (under polynomial reductions). On the other hand the problem is probably not NP-Complete. Recently, Fredman and Khachiyan (1994) have presented a sub-exponential $n^{O(\log n)}$ time algorithm for the HTR problem.

We first show that the problem CCM is at least as hard as HTR. By that we mean that if there is an output polynomial algorithm for CCM then there is an output polynomial algorithm for HTR. This has been stated as an open problem by Kavvadias et. al. (1993), who proved a similar hardness result for SID. Both hardness results can be alternatively derived by combining previous results in database theory (Eiter & Gottlob, 1994; Bioch & Ibaraki, 1993) and its relation to our problems (Khardon et al., 1995).

We then consider two relaxations of these translation problems. The first is considering redundant Horn expressions which contain *all* the Horn prime implicates for a given expression. The output of SID is therefore altered to be the set of all prime implicates, and similarly the input of CCM includes all the prime implicates instead of a minimal subset. It is shown that in this special case, SID, CCM, and HTR are equivalent under polynomial reductions. Therefore, the algorithm presented by Fredman and Khachiyan (1994) can be used to solve CCM, and SID in time $n^{O(\log n)}$. This result can be alternatively derived from the results on functional dependencies in MAK form (Eiter & Gottlob, 1991). We show however that our argument generalizes to the larger family of $k$-quasi Horn expressions.

The second relaxation is the problem of computing all the prime implicants for a given Horn expression. This is a relaxation of CCM since using the prime implicants one can compute the characteristic models. Interestingly, the algorithm for HTR (Fredman & Khachiyan, 1994) can be adapted to this problem, resulting an algorithm with time complexity $n^{O(\log^2 n)}$.

It is shown, however, that both relaxations do not help in solving the general cases of CCM and SID due to exponential gaps in the size of the corresponding representations.

Lastly, we consider a related problem, denoted EOC, which is a minor modification of CCM and SID. This problem is shown to be co-NP-Complete. This serves to highlight some of the difficulty in finding the exact complexity of our problems. A variant of this result, has already appeared in the database literature (Gottlob & Libkin, 1990).

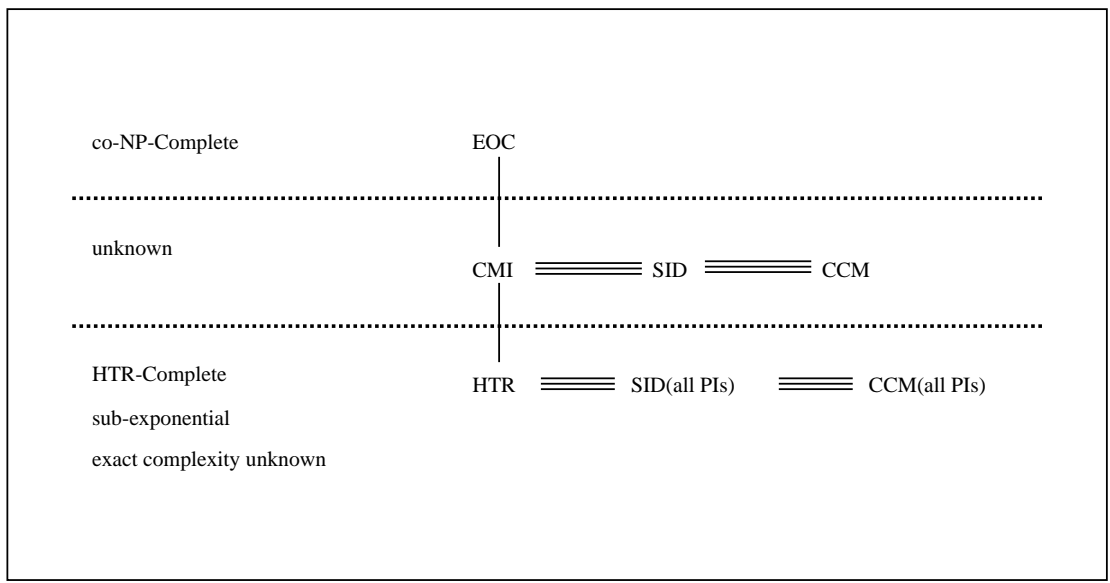

Figure 1: Summary of Complexity Results

Our results are summarized in Figure 1, where a hierarchy of problems is depicted. The problem EOC is co-NP-Complete. The problem CMI is a special case of EOC, and is equivalent to SID and CCM. The problem HTR is a special case of CMI and is equivalent to SID and CCM under the restriction that the Horn expression is represented by the set of all prime implicates.

The rest of the paper is organized as follows. Section 2 defines characteristic models, describes some of their properties, and formally defines the problems in question. Section 3 discusses the relation between CCM,SID and the corresponding decision problem. Section 4 discusses the relation to the HTR problem. We first establish the hardness result, and then consider the two relaxations mentioned above. Section 5 shows that EOC is co-NP-Hard, and Section 6 concludes with a summary.

## 2. Preliminaries

This section includes the basic definitions, and introduces several previous results which are used in the paper.

We consider Boolean functions $f : \{0,1\}^n \rightarrow \{0,1\}$. The elements in the set $\{x_1, \ldots, x_n\}$ are called variables. Assignments in $\{0,1\}^n$ are denoted by $x, y, z$, and $weight(x)$ denotes the number of 1 bits in the assignment $x$. A literal is either a variable $x_i$ (called a positive literal) or its negation $\overline{x_i}$ (a negative literal). A clause is a disjunction of literals, and a CNF formula is a conjunction of clauses. For example $(x_1 \vee \overline{x_2}) \wedge (x_3 \vee \overline{x_1} \vee x_4)$ is a CNF formula with two clauses. A term is a conjunction of literals, and a DNF formula is a disjunction of terms. For example $(x_1 \wedge \overline{x_2}) \vee (x_3 \wedge \overline{x_1} \wedge x_4)$ is a DNF formula with two terms. A CNF formula is Horn if every clause in it has at most one positive literal. A formula is monotone if all the literals that appear in it are positive. The size of CNF and DNF representations is, respectively, the number of clauses and the number of terms in the representation. We denote by $|DNF(f)|$ the size of the smallest DNF representation for $f$.

An assignment $x \in \{0,1\}^n$ satisfies $f$ if $f(x) = 1$. Such an assignment $x$ is also called a model of $f$. By "$f$ implies $g$", denoted $f \models g$, we mean that every model of $f$ is also a model of $g$. Throughout the paper, when no confusion can arise, we identify a Boolean function $f$ with the set of its models, namely $f^{-1}(1)$. Observe that the connective "implies" ($\models$) used between Boolean functions is equivalent to the connective "subset or equal" ($\subseteq$) used for subsets of $\{0,1\}^n$. That is, $f \models g$ if and only if $f \subseteq g$.

A term $t$ is an *implicant* of a function $f$, if $t \models f$. A term $t$ is a *prime implicant* of a function $f$, if $t$ is an implicant of $f$ and the conjunction of any proper subset of the literals in $t$ is not an implicant of $f$.

A clause $d$ is an *implicate* of a function $f$, if $f \models d$. A clause $d$ is a *prime implicate* of a function $f$, if $d$ is an implicate of $f$ and the disjunction of any proper subset of the literals in $d$ is not an implicate of $f$.

It is well known that, a minimal DNF representation of $f$ is a disjunction of some of its prime implicants. A minimal CNF representation of $f$ is a conjunction of some of its prime implicates.

If $f$ is monotone, then it has a unique minimal DNF representation (using all the prime implicants), and a unique minimal CNF representation (using all its prime implicates).

## 2.1 Characteristic Models

The idea of using characteristic models as a knowledge representation was introduced by Kautz et. al. (1995). Characteristic models were studied in AI (Dechter & Pearl, 1992; Kavvadias et al., 1993; Khardon & Roth, 1994) and under a different manifestation in database theory (Beeri et al., 1984; Mannila & Raiha, 1986; Gottlob & Libkin, 1990; Eiter & Gottlob, 1991, 1994). This section defines characteristic models and their basic properties.

For $u, v \in \{0,1\}^n$, we define the *intersection* of $u$ and $v$ to be the assignment $z \in \{0,1\}^n$ such that $z_i = 1$ if and only if $u_i = 1$ and $v_i = 1$ (i.e., the bitwise logical-and of $u$ and $v$.).

For a set of assignments $S$, $x = intersect(S)$ is the assignment we get by intersecting all the assignments in $S$. We say that $S$ is *redundant* if there exists $x \in S$ and $S' \subseteq S$ such that $x \notin S'$ and $x = intersect(S')$. Otherwise $S$ is *non-redundant*.

The *closure* of $S \subseteq \{0,1\}^n$, denoted $closure(S)$, is defined as the smallest set containing $S$ that is closed under intersection.

To illustrate these definitions consider the set $M = \{1101, 1110, 0101\}$. Then $M$ is non-redundant, $intersect(M) = 0100$, and $closure(M) = \{1101, 1110, 0101, 0100, 1100\}$.

Let $H$ be a Horn expression. The set of the *Horn characteristic models* of $H$, denoted here $char(H)$ is defined as the set of models of $H$ that are not the intersection of other models of $H$. Note that $char(H)$ is non-redundant. Formally,

$$char(H) = \{u \in H \mid u \notin closure(H \setminus \{u\})\}. \tag{1}$$

For example, $char(\{1101, 1110, 0101, 0100\}) = \{1101, 1110, 0101\}$.

It is well known that the set of models of Horn expressions is closed under intersection. This result is due to McKinsey (1943), who proved it for a certain class of first order sentences. Alfred Horn (1951) considered a more general class of sentences. (Lemma 7 by Horn (1951) deals with the propositional case. Dechter and Pearl (1992) present another proof for the propositional case.) Moreover, since characteristic models capture all the information about the closure, they also capture all the information about the Horn expression.

**Theorem 1 (Kautz et al., 1995; Dechter & Pearl, 1992)** *Let $H$ be a Horn expression then $H = closure(char(H))$.*

## 2.2 Monotone Theory and Characteristic Models

The monotone theory was introduced by Bshouty (1993), and was later used for a theory for model-based reasoning (Khardon & Roth, 1994). This section explores the relations between the monotone theory and characteristic models.

**Definition 1 (Order)** *We denote by $\leq$ the* usual partial order *on the lattice $\{0,1\}^n$, the one induced by the order $0 < 1$. That is, for $x, y \in \{0,1\}^n$, $x \leq y$ if and only if $\forall i, x_i \leq y_i$. For an assignment $b \in \{0,1\}^n$ we define $x \leq_b y$ if and only if $x \oplus b \leq y \oplus b$ (Here $\oplus$ is the bitwise addition modulo 2). We say that $x > y$ if and only if $x \geq y$ and $x \neq y$.*

Intuitively, if $b_i = 0$ then the order relation on the $i$th bit is the normal order; if $b_i = 1$, the order relation is reversed and we have that $1 <_{b_i} 0$. For example $0101 <_{1111} 0100$, and $0101 \not<_{1111} 0110$. We now define:

The *monotone extension of $z \in \{0,1\}^n$* with respect to $b$:

$$\mathcal{M}_b(z) = \{x \mid x \geq_b z\}.$$

The *monotone extension of $f$* with respect to $b$:

$$\mathcal{M}_b(f) = \{x \mid x \geq_b z, \text{ for some } z \in f\}.$$

The set of *minimal assignments of $f$* with respect to $b$:

$$\min_b(f) = \{z \mid z \in f, \text{ such that } \forall y \in f, z \not>_b y\}.$$

For example

$$\mathcal{M}_{1111}(0101) = \{0101, 0001, 0100, 0000\}, \quad \text{and}$$

$$\mathcal{M}_{1111}(1100) = \{1100, 0100, 1000, 0000\}.$$

Let $f = b\overline{c}(\overline{a} \vee \overline{d})(a \vee d)$, then in the set notation $f = \{1100, 0101\}$, and $\mathcal{M}_{1111}(f) = \{0101, 0001, 0100, 0000, 1100, 1000\}$. The set $\min_{1111}(f) = \{1100, 0101\}$, and the set $\min_{0001}(f) = \{0101\}$.

Clearly, for every assignment $b \in \{0,1\}^n$, $f \subseteq \mathcal{M}_b(f)$. Moreover, if $b \notin f$, then $b \notin \mathcal{M}_b(f)$ (since $b$ is the smallest assignment with respect to the order $\leq_b$). Therefore:

$$f = \bigwedge_{b \in \{0,1\}^n} \mathcal{M}_b(f) = \bigwedge_{b \notin f} \mathcal{M}_b(f).$$

The question is if we can find a small set of negative examples, and use it to represent $f$ as above.

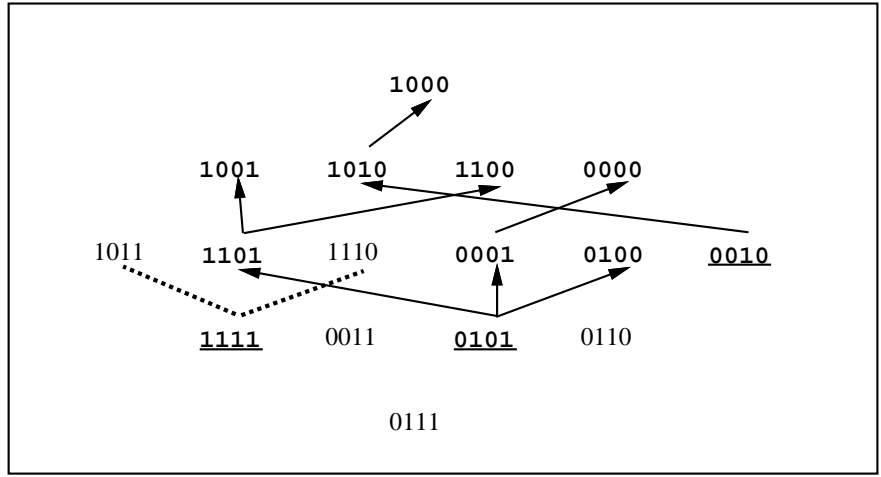

Figure 2: Computing $\min_b(f)$ and $\mathcal{M}_b(f)$

**Definition 2 (Basis)** *A set $B$ is a basis for $f$ if $f = \bigwedge_{b \in B} \mathcal{M}_b(f)$. $B$ is a basis for a class of functions $\mathcal{F}$ if it is a basis for all the functions in $\mathcal{F}$.*

Using this definition, we get an alternative representation for functions

$$f = \bigwedge_{b \in B} \mathcal{M}_b(f) = \bigwedge_{b \in B} \bigvee_{z \in \min_b(f)} \mathcal{M}_b(z). \qquad (2)$$

It is known that the set $B_H = \{u \in \{0,1\}^n \mid weight(u) \geq n-1\}$, is a basis for any Horn CNF function. For example consider the Horn expression $W = (bc \rightarrow d)(cd \rightarrow b)(bc \rightarrow a)$ discussed in the introduction. Recall that the satisfying assignments of $W$ are:

$$models(W) = \{0000, 0001, 0010, 0100, 0101, 1000, 1001, 1010, 1100, 1101, 1111\}.$$

We have to compute the sets $\min_b(W)$ for $b \in B_H$, where $B_H = \{1111, 1110, 1101, 1011, 0111\}$. Note that if $b$ satisfies $f$ then $\min_b(f) = \{b\}$, and $\mathcal{M}_b(f) \equiv 1$ (that is, $\forall x$, $\mathcal{M}_b(f)(x) = 1$). Therefore, $\min_{1111}(W) = \{1111\}$, and $\min_{1101}(W) = \{1101\}$. One way to compute the sets of minimal assignments is by drawing the corresponding lattices and noting the relations there. Figure 2 shows the lattice with respect to $b = 0111$. The satisfying assignments of $W$ are marked in bold face. The minimal assignments are underlined, and some of the order relations, which show that the rest of the assignments are not minimal, are drawn. To compute $\mathcal{M}_b(W)$ we have to add any assignment which is above the minimal assignments. This is marked by the dotted lines which show that $1011$ and $1110$ are in $\mathcal{M}_{0111}(W)$. Using the figure we observe that $\min_{0111}(W) = \{1111, 0101, 0010\}$. The other sets are $\min_{1110}(W) = \{1111, 1100, 1010\}$, and $\min_{1011}(W) = \{1111, 1001, 1010\}$.

It is known that the size of the basis for a function $f$ is bounded by the size of its CNF representation, and that for every $b$ the size of $\min_b(f)$ is bounded by the size of its DNF representation.

For any function $f$ and set of assignments $B$ let:

$$\Gamma_f^B = \min_B(f) = \cup_{b \in B} \{z \in \min_b(f)\}.$$

The following theorem gives an alternative way to define $char(H)$.

**Theorem 2 (Khardon & Roth, 1994)** *Let $H$ be a Horn expression. Then $char(H) = \Gamma_H^{B_H}$.*

Continuing the above example with the function $W = (bc \rightarrow d)(cd \rightarrow b)(bc \rightarrow a)$, we conclude that $char(W) = \{0010, 0101, 1001, 1010, 1100, 1101, 1111\}$. As the following theorem shows the set of characteristic models can be used to answer deduction queries.

**Theorem 3 (Kautz et al., 1995; Khardon & Roth, 1994)** *Let $H_1$, $H_2$ be Horn expressions then $H_1 \models H_2$ if and only if for all $x \in char(H_1)$, $H_2(x) = 1$.*

It is useful to have the DNF representation of a function. If $f$ is given in its DNF representation then it is easy to compute the set $min_b(f)$, for any $b$. Each term in the DNF representation can contribute *at most* one assignment, $min_b(t)$, where the variables that appear in the term are fixed and the others are set to their minimal value. This is true since from every other satisfying assignment of the term we can "walk down the lattice" towards this assignment, on a path composed of satisfying assignments. For example, the minimal assignment for the term $t = x_1\overline{x_3}$, with respect to the basis element $b = 0011$, is $min_{0011}(t) = \{1001\}$. The assignment $1100$ which also satisfies $t$ is not minimal since $1001 <_{0011} 1101 <_{0011} 1100$. Further, once we have one assignment from each term, it is easy make sure that the set is non-redundant by checking which of the assignments generated is in the intersection of the others. We would use this algorithm later in some of our reductions.

We say that a function is $b$-monotone if it is monotone according to the order relation $\leq_b$. Namely, if whenever $f(x) = 1$ and $y \geq_b x$ then $f(y) = 1$. Notice that if we rename the variable $x_i$ by its negation, for each $i$ such that $b_i = 1$ (i.e. where the order relation is reversed), then $f$ becomes monotone. Therefore, $b$-monotone functions enjoy similar properties. For example, they have unique minimal DNF and CNF representations. Another property is that the minimal assignment which corresponds to every term is indeed part of the set $min_b(f)$.

**Claim 1 (Khardon & Roth, 1994)** *For any $b$-monotone function $f$, there is a 1-1 correspondence between the prime implicants of $f$ and the set $min_b(f)$. Namely:*
*(1) for every term $t$ in the minimal DNF representation for $f$, the assignment $min_b(t)$ is in $min_b(f)$.*
*(2) $|min_b(f)| = |DNF(f)|$.*

We would also use the notion of a least upper bound of a Boolean function (Selman & Kautz, 1991), which can sometimes be characterized by the monotone theory.

**Definition 3 (Least Upper-bound)** *Let $\mathcal{F}, \mathcal{G}$ be classes of Boolean functions. Given $f \in \mathcal{F}$ we say that $g \in \mathcal{G}$ is a $\mathcal{G}$-least upper bound of $f$ if and only if $f \subseteq g$ and there is no $f' \in \mathcal{G}$ such that $f \subseteq f' \subset g$.*

**Theorem 4 (Khardon & Roth, 1994)** *Let $f$ be any Boolean function and $\mathcal{G}$ a class of all Boolean functions with basis $B$. Then, $f_{lub}^{B}$ defined as*

$$f_{lub}^{B} = \bigwedge_{b \in B} \mathcal{M}_b(f)$$

*is the $\mathcal{G}$-least upper bound of $f$.*

For the class of Horn expressions we have two ways to express the least upper bound. One using the monotone theory, and one using the *closure* operator:

**Theorem 5 (Dechter & Pearl, 1992; Kautz et al., 1995; Khardon & Roth, 1994)** *Let $f : \{0,1\}^n \to \{0,1\}$ be a Boolean function. Then $f_{lub}^{B_H} = closure(f)$, and $char(f_{lub}^{B_H}) \subseteq f$.*

For example consider the function $f = (bc \to d)(cd \to b)(bc \to a)(a \vee b \vee c \vee \overline{d})$. The function $f$ satisfies all the assignments as $W$ above except for 0001. However, $intersect(\{0101, 1001\}) = 0001$, and therefore $f_{lub}^{B_H} = W$.

## 2.3 The Computational Problems

This section includes definitions for all the problems discussed in this paper. Let $H$ be a CNF expression in Horn form, and let $char(H)$ be its set of characteristic models. The translation problems considered are:

**CCM:** Computing Characteristic Models
Input: a Horn CNF $H$.
Output: the set $char(H)$.

**SID:** Structure Identification (Computing Horn Expressions)
Input: a set of assignments $\Gamma$.
Output: a Horn CNF $H$, such that $\Gamma = char(H)$.

**HTR:** Hypergraph Transversals (Dualization of Monotone Expressions)
Input: a monotone CNF expression $C$.
Output: a monotone DNF expression $D$, such that $C \equiv D$.

The decision problems discussed:

**CMI:** Characteristic Models Identification
Input: a Horn CNF $H$, and a set $G$ of satisfying assignments of $H$.
Output: Yes iff $char(H) \subseteq G$.
Note: The condition is equivalent to $H \models closure(G)$, and essentially also to $G = char(H)$.

**EOC:** Entailment of Closure
Input: a Horn CNF $H$, a set $G$ of assignments.
Output: Yes if and only if $H \models closure(G)$.

We also discuss the following variant of CMI:

**CMIC:** Characteristic Models Identification with Counter example
Input: a Horn CNF $H$, a set $G$ of satisfying assignments of $H$.
Output: If $Char(H) \subseteq G$ then output Yes. Otherwise, output No and supply a counter example $x \in Char(H) \setminus G$.

### 2.4 Polynomial Time Algorithms and Reductions

As mentioned above we need to define algorithms that are polynomial with respect to their output. There is more than one way to give such a definition. (A discussion of this issue is given by Eiter and Gottlob (1994).) We use the weakest[1] of those which is called an *output polynomial* algorithm.

When the output of a problem $P$ is uniquely defined, we say that an algorithm $A$ is an output polynomial algorithm for $P$ if it solves $P$ correctly in time which is polynomial in the size of its input and output. This is the case with HTR, and CCM.

When the output of a problem $P$ is not uniquely defined, we consider the shortest permissible output $O(I)$ for input $I$. We say that an algorithm $A$ is an output polynomial algorithm for $P$ if it solves $P$ correctly in time which is polynomial in the size of its input $I$ and the size of $O(I)$. We note that for SID the output is not uniquely defined since there is no unique minimal representation for Horn functions.

We define polynomial reductions with respect to an oracle (i.e. we use Turing reducibility (Garey & Johnson, 1979)). A problem P1 is *polynomially reducible* to a problem P2 if there is an output polynomial algorithm that solves P1 when given access to (1) an output polynomial subroutine for P2, and (2) a polynomial bound[2] on the running time of the subroutine.

## 3. Translating is Equivalent to Deciding

In this section we show that the problems CCM, SID, CMI, and CMIC are equivalent under polynomial reductions. Namely, both translation problems are solvable in polynomial time if and only if the corresponding decision problem CMI is solvable in polynomial time.

**Theorem 6** *The problems CCM,SID,CMI, and CMIC are equivalent under polynomial reductions.*

**Proof:** The proof is established in a series of lemmas. In particular we show that CMIC $\leq$ CMI $\leq$ SID $\leq$ CMIC, and that CMI $\leq$ CCM $\leq$ CMIC, where $\leq$ denotes "is polynomially reducible to", in Lemma 1, Lemma 2, Lemma 3, Lemma 4, and Lemma 5 respectively. ■

**Lemma 1** *The problem CMIC is polynomially reducible to the problem CMI.*

Before presenting the proof consider how a similar result is achieved for the satisfiability problem (Garey & Johnson, 1979). Namely, how a decision procedure for satisfiability can be used to construct an algorithm that finds a satisfying assignment if one exists. Suppose we have a formula $C$, and that we know that it is satisfiable. (We used the decision procedure to find that out.) Our task is to find a satisfying assignment for it. What we do is substitute $x_1 = 0$ into $C$ yielding a formula $C^0$ with $n - 1$ variables. The formula $C^0$ is satisfiable if and only if $C$ has a satisfying assignment in which $x_i = 0$. We run the decision procedure on $C^0$. If the answer is Yes then we know that $C$ has a satisfying assignment in which $x_i = 0$. If the answer is No then since $C$ is satisfiable, it must have a satisfying assignment in which $x_1 = 1$. In either case we found a substitution for $x_1$ which guarantees the existence of a satisfying assignment. All we have to do is to recurse with this procedure on $C^0$.

---

1. Other related notions which we do not use here are "enumeration with polynomial delay" and "enumeration with incremental polynomial delay" (Eiter & Gottlob, 1994). These require that the algorithm will compute the elements of its output one at a time, and restrict the time delay between consecutive outputs. Incremental polynomial delay allows the delay to depend on the problem size and on the number of elements computed so far. Polynomial delay is stricter in that it requires dependence only on the problem size. Both of these notions are stricter than output polynomial algorithms since the latter may wait a long time before computing its first output. Unfortunately, most of our reductions yield output polynomial algorithms, and we cannot guarantee that the stronger notions hold.

2. That is, a polynomial in the dimension of the problem (the number of variables), the input size, and the output size.

An example can clarify this a bit more. Suppose we have the expression $C = (a \vee c)(b \vee \overline{c})$ which is satisfiable. To find a satisfying assignment we substitute $a = 0$ to get $C^0 = (c)(b \vee \overline{c})$, and run the decision procedure on $C^0$. The answer is Yes, and therefore we continue with $C^0$. We next substitute $b = 0$ to get $C^{00} = c\overline{c}$. We run the decision procedure again, and the answer is No. Therefore we conclude that we must substitute $b = 1$ instead of $b = 0$. This yields $C^{01} = c$. We then continue to find that $c$ must be assigned 1 and altogether we find the satisfying assignment $abc = 011$.

We would like to use the same trick here. However, $G$ is given as a set of models and we cannot perform this substitution procedure as easily[3]. Nevertheless, as the proof shows something similar can be done.

**Proof:** First observe that we have a solver for CMI. Therefore if the answer is Yes we have no problem, we can simply answer Yes. A problem arises in the case where the answer is No. In this case Many is happy with saying No, but CMIC must provide a counter example.

Formally, we get $H, G$ as input to CMIC and an algorithm $A$ to solve CMI. We run $A$ on $H, G$ as an input, and if $A$ replies Yes we reply Yes. Otherwise we know that there exits an $x \in char(H) \setminus G$. We need to find such a model and return it as the output of CMIC.

Consider first the easier task of finding $x \in H \setminus closure(G)$; the assignment $x$ is a *witness* for the fact $H \not\models closure(G)$.

Recall the substitution trick from above, and observe that for $x_i = 1$ a similar substitution works. For $H$ we simply perform the substitution to get an expression $\tilde{H}$, and for $G$ we remove any $z \in G$ in which $z_i = 0$ to get the set $\tilde{G}$. We claim that there is a witness for $H, G$ with $x_i = 1$ if and only if there is a witness for $\tilde{H}, \tilde{G}$. This follows from the fact that $x \in closure(G)$ and $x_i = 1$ if and only if $x \in closure(\tilde{G})$. To see that, let $x \in closure(G)$, such that $x_i = 1$; if $x = intersect(S)$, and $y \in S$ then $y_i = 1$, and therefore $x \in closure(\tilde{G})$. Also if $x \in closure(\tilde{G})$ then $x \in closure(G)$. Therefore, if there is a witness $x$ with $x_i = 1$ then we can detect this fact by presenting $A$ with $\tilde{H}, \tilde{G}$ as input (on which it will say No).

This however does not work for $x_i = 0$. In this case an element in the closure requires at least one element in $S$ with $y_i = 0$, but we have no information on the other elements. Therefore we can not perform the recursion in the case where substitution of $x_i = 0$ is required.

We circumvent this problem using the following iterative procedure. In each stage we try to turn one more variable to 1. For all $i$, we make the experiment described above of substituting $x_i = 1$. If the answer is No, for some $i$, we can proceed to the next stage, just as before (ignoring tests for other values of $i$). If the answer is Yes for all $i$, then we know that for each $x_i$ that did not receive a value so far, there is no witness with $x_i = 1$, so the only possible witness is the one assigning 0 to all the variables. We return the witness $x \in \{0, 1\}^n$ arrived at, by the above substitutions, as the counter example of CMIC.

---

3. Furthermore, the closed form one can derive using the monotone theory (Khardon & Roth, 1994) does not seem to be useful.

From the construction it is clear that $x \in H \setminus closure(G)$, but the requirement of CMIC is that $x \in char(H) \setminus G$. We claim that this stronger condition holds. Suppose not, and let $S \subseteq char(H)$ be such that $x = intersect(S)$. Then clearly $S$ is not a subset of $G$ or otherwise $x \in closure(G)$. Let $y \in S \setminus G$, then since $x = intersect(S)$, we get $x <_{0^n} y$. Namely, if $x_i = 1$ then $y_i = 1$. But this is a contradiction, since in the last run of the algorithm $A$ for CMI, it was concluded that no more variables could be set to 1, while still maintaining a witness. ∎

We exemplify the proof using the function $W = (bc \rightarrow d)(cd \rightarrow b)(bc \rightarrow a)$ presented in the introduction. Recall that $char(W) = \{0010, 0101, 1001, 1010, 1100, 1101, 1111\}$, and suppose that so far we found $G = \{0010, 1001, 1010, 1100, 1101, 1111\}$. That is, all but the model 0101. We run CMI on $W, G$ and, since $G$ does not include all the characteristic models, it answers No. In order to find the counter example we make 4 separate substitutions, one for each variable substituted to 1.

Consider the substitution with $b = 1$. This yields $\tilde{W} = (c \rightarrow d)(c \rightarrow a)$, and $\tilde{G} = \{1s00, 1s01, 1s11\}$, where we use $s$ to mark that the variable $b$ was substituted. We run CMI on $\tilde{W}, \tilde{G}$ and it finds out that there is a counter example (the assignment $0s01$ is in $\tilde{W}$ but not in $closure(\tilde{G})$), and therefore it answers No. That means we can continue our algorithm with $b = 1$. We forget all the information from the other substitutions (that were not considered in detail) and continue to the next step.

In the next step we substitute 1 to each of $a, c, d$. Consider first the substitution for $a$. This yields $\tilde{W} = (c \rightarrow d)$ and $\tilde{G} = \{ss00, ss01, ss11\}$. Running CMI on this pair we get the answer Yes. Namely $\tilde{W} = closure(\tilde{G})$. Consider now the substitution for $d$. This yields $\tilde{W} = (c \rightarrow a)$ and $\tilde{G} = \{1s0s, 1s1s\}$. Running CMI on this pair we get the answer No (since $0s0s$ is a counter example). We can therefore recurse on this value.

In the next iteration both substitutions for $a$ and for $c$, yield the answer Yes, and therefore we substitute 0 to both to get the final counter example $abcd = 0101$.

Using this example it is easy to see that one can improve the running time of the reduction by simply remembering the attributes for which we received the answer Yes. These attributes will have to get the value 0 in the end. In this way we can scan the variables one by one, and recurse on the first that yields the answer No. This requires only $n$ calls to CMI.

**Lemma 2** *The problem CMI is polynomially reducible to the problem SID.*

**Proof:** We are given an output polynomial time algorithm $A$ for SID, and a polynomial bound on its running time (that is, a polynomial in the number of variables $n$, the input size, and the output size). Given $H, G$ as input to CMI, we run $A$ on $G$ until it stops and outputs $H'$ or until it exceeds its time bound (with respect to the size of $H$). In the first case we check whether $H = H'$ (which can be done in polynomial time (Dowling & Gallier, 1984)) and answer accordingly. In the second case we know that the real Horn expression which corresponds to $G$ is larger than $H$ and therefore we answer No. ∎

The proof of the next lemma draws on previous results in computational learning theory. In this framework a function $f : \{0,1\}^n \rightarrow \{0,1\}$ is hidden from a learner that has to reproduce it by accessing certain "oracles". A membership query allows the learner to ask for the value of the function on a certain point.

**Definition 4** *A membership query oracle for a function $f : \{0,1\}^n \rightarrow \{0,1\}$, denoted $MQ(f)$, is an oracle that when presented with $x \in \{0,1\}^n$ returns $f(x)$.*

An equivalence query allows the learner to find out whether a hypothesis he has is equivalent to $f$ or not. In case it is not equivalent, the learner is supplied with a counter example.

**Definition 5** *An equivalence query oracle for a function $f : \{0,1\}^n \rightarrow \{0,1\}$, denoted $EQ(f)$, is an oracle that when presented with a hypothesis $h : \{0,1\}^n \rightarrow \{0,1\}$, returns Yes if $f \equiv h$. Otherwise it returns No and a counter example $x$ such that $f(x) \neq h(x)$.*

We use a result that has been obtained in this framework.

**Theorem 7 (Angluin, Frazier, & Pitt, 1992)** *There is an algorithm $A$, that when given access to $MQ(f)$ and $EQ(f)$, where $f$ is a hidden Horn expression, runs in time polynomial in the number of variables and in the size of $f$, and outputs a Horn expression $H$ which is equivalent to $f$.*
*The hypothesis $h$, in the algorithm's accesses to $EQ(f)$, is always a Horn expression.*

The following lemma, and the simulation in its proof, are implicit in previous works (Dechter & Pearl, 1992; Kautz et al., 1995; Kivinen & Mannila, 1994).

**Lemma 3** *The problem SID is polynomially reducible to the problem CMIC.*

**Proof:** We are given $G$ as input to SID, and a polynomial time algorithm $C$ for CMIC. Our algorithm will run the algorithm $A$ from Theorem 7 and answer the $MQ$ and $EQ$ queries that $A$ presents.

Given $x \in \{0,1\}^n$ for $MQ$ the algorithm tests whether $x \in closure(G)$. This can be done by testing whether $x$ is equal to the intersection of all elements $y$ in $G$ such that $y \geq x$.

Given a Horn expression $h$ for $EQ$ (the theorem guarantees that the hypothesis is a Horn expression), we have to test whether $h \equiv closure(G)$. We first test whether $closure(G) \subseteq h$, which is equivalent to $closure(G) \models h$. Theorem 5 together with Theorem 3 imply that if the answer is No, then for some $x \in G$, $h(x) = 0$. Such an $x$ is a counter example for the equivalence query, and the test can be performed simply by evaluating $h$ on all the assignments in $G$.

If $closure(G) \models h$, namely all the assignments in $G$ satisfy $h$, we present $h, G$ as input to the algorithm $C$ for the problem CMIC. The input to CMIC is legal. $C$ may answer Yes, meaning $char(h) \subseteq G$, which implies $h \subseteq closure(G)$. In this case we answer Yes to the equivalence query. Otherwise $C$ says No and supplies a counter example $x \in char(h) \setminus G$. Since $G \subseteq h$ we get $x \in h \setminus closure(G)$ and therefore we can pass $x$ on as a counter example to the equivalence query. ∎

We next consider the problem CCM:

**Lemma 4** *The problem CMI is polynomially reducible to the problem CCM.*

**Proof:** We are given an output polynomial algorithm $C$ for CCM, and a polynomial bound on its running time (that is, a polynomial in the number of variables $n$, the input size, and the output size). Given $H, G$ as input to CMI, we run $C$ on $H$ until it stops and outputs $G'$ or until it exceeds its time bound (with respect to the size of $G$). In the first case we compare $G$ and $G'$ and answer accordingly. In the second case we know that the set of characteristic models of $H$ is larger than $G$ and therefore we answer No. ∎

**Lemma 5** *The problem CCM is polynomially reducible to the problem CMIC.*

**Proof:** Given $H$ as input for CCM, an algorithm for CMIC can be used repeatedly to produce the elements of $char(H)$.

We start with $G = \emptyset$. In each iteration we run CMIC on $H, G$ to get a new characteristic model which we add to $G$. Once we find all the characteristic models CMIC will answer Yes. (In fact, if CMIC is polynomial in its input size then we get an "incremental polynomial algorithm" (Eiter & Gottlob, 1994) which is even stronger than "output polynomial" as required here.) ∎

## 4. The Relation to Hypergraph Transversals

In this section we establish the relation to the hypergraph transversal problem. We first show that our problems are at least as hard as HTR. We then consider two relaxations of SID and CCM. The first relaxation considers redundant representation for Horn expressions, which includes all the prime implicates. The second relaxation considers computing prime implicants instead of characteristic models. Both of these relaxations enjoy sub-exponential algorithms. It is shown, however, that the relaxations do not help in the general case, as a result of exponential gap in the size of the corresponding representations.

### 4.1 The Reduction to HTR

The problem HTR is defined as computing a DNF representation for a monotone function given in its CNF form. It is easy to observe that this is equivalent to computing a CNF representation for a monotone function given in its DNF form. (We can simply exchange the $\vee$ and $\wedge$ operations to get one problem from the other). We can therefore assume that the input for HTR is given as either a DNF or a CNF. Another useful observation is that renaming the variables does not change the problem. Therefore if we rename every variable as its negation (namely, replace every $x_i$ with $\overline{x_i}$), we get the equivalent problem of translating between functions which are monotone with respect to the order relation $\leq_{1^n}$. We call such functions *anti-monotone*. This is useful since anti-monotone functions have CNF representations in which all variables are negated, which is a special case of Horn expressions. Having these observations, the next two theorems follow almost immediately from the definitions, given the correspondence between minimal elements and prime implicants described in Claim 1. The following result has been stated as an open problem by Kavvadias et. al. (1993).

**Theorem 8** *The problem HTR is polynomially reducible to the problem CCM.*

**Proof:** Let $A$ be an algorithm for the problem CCM. We construct an algorithm $B$ for the problem HTR. We may assume that the input is an anti-monotone CNF, $C$, and we want to compute its anti-monotone DNF representation.

The basic idea is that using Claim 1 we know how to compute the DNF from $\min_{1^n}(C)$, and that the latter is a subset of the characteristic models. So all we need to do is let $A$ compute the characteristic models, identify the set $\min_{1^n}(C)$, and compute the DNF.

More formally, the algorithm $B$ runs $A$ to compute $\Gamma = char(C) = \min_{B_H}(C)$, and computes the set $\Gamma_{1^n} = \{z \in \Gamma \mid \forall y \in \Gamma, z \not\prec_{1^n} y\}$. Namely the elements of $\Gamma$ which are minimal with respect to the order relation $b = 1^n$. It then computes the anti-monotone DNF expression $D = \vee_{z \in \Gamma_{1^n}} \wedge_{z_i = 0} \overline{x_i}$, which it outputs.

The correctness of the algorithm follows from Claim 1 which guarantees that the computation of the DNF from the set of characteristic models is correct.

As for the time complexity we observe, using Claim 1, that $\Gamma$ is not considerably larger than the size of the DNF. This is true since for all $b$, $|DNF(f)| = |\min_{1^n}(f)| \ge |\min_b(f)|$, and $|B_H| = n + 1$. ∎

To exemplify the above reduction, suppose that we have only three variables $a, b, c$, and that the input is $C = (\overline{a} \vee \overline{b})(\overline{b} \vee \overline{c})$. (The satisfying assignments are $000, 001, 010, 100, 101$, and the required DNF expression is $\overline{a} \; \overline{c} \vee \overline{b}$.) The algorithm $A$ will compute the set of characteristic models $char(C) = \{101, 010, 100, 001\}$, from that we find that $min_{1^n}(C) = \{101, 010\}$. The term which corresponds to $101$ is $\overline{b}$, and the term which corresponds to $010$ is $\overline{a} \; \overline{c}$ and indeed we get the right DNF expression.

Using the monotone theory one can give a simple proof for the following theorem, which has already been proved by Kavvadias et. al. (1993).

**Theorem 9 (Kavvadias et al., 1993)** *The problem HTR is polynomially reducible to the problem SID.*

We note that both theorems can be deduced by combining results in database theory (Eiter & Gottlob, 1994, 1991; Bioch & Ibaraki, 1993) and using the above mentioned equivalence with problems in database theory (Khardon et al., 1995).

## 4.2 Enumerating Prime Implicates

Having obtained the hardness results in the previous sub-section, a natural question is whether CCM, and SID are as easy as HTR. This would help settle the exact complexity of the problems discussed, and more importantly would imply a sub-exponential algorithm for the problem. While no such reduction has been found, we show here that it holds in a special case. We show, however, that the solution obtained in this way may need exponential time in the general case.

This result has already been obtained in the database domain (Eiter & Gottlob, 1991), where restrictions of functional dependencies to be in MAK form is discussed. Our argument, however, can be generalized to richer languages, and in particular holds for the family of $k$-quasi Horn expressions defined below.

In particular we relax the problems so as to use the largest Horn expression for a function instead of using a small Horn expression. In this case the problem SID amounts to computing all the (Horn) prime implicates of the function identified by $\Gamma$. For CCM we have to compute the set of characteristic models given the set of all prime implicates rather than a small expression.

We would use the following example to illustrate the notions in this sub-section. Consider the function $W = (a \rightarrow b)(c \rightarrow b)(\overline{b} \vee \overline{d})$. The satisfying assignments of $W$ are $W = \{0000, 0001, 0100, 0110, 1100, 1110\}$, and the characteristic models are $char(W) = \{0001, 0110, 1100, 1110\}$. One can verify that $W \models (\overline{c} \vee \overline{d})(\overline{a} \vee \overline{d})$, and that these are the only additional Horn prime implicates of $W$.

For CCM, this section asks whether it is easier to compute the characteristic models starting with the equivalent expression $W = (a \rightarrow b)(c \rightarrow b)(\overline{b} \vee \overline{d})(\overline{c} \vee \overline{d})(\overline{a} \vee \overline{d})$. For SID the question is whether it is easier to output the whole set rather than just a minimal subset. These are relaxations of the problems since, an algorithm for SID is allowed more time to compute its output, and CCM is given more information and more time for its computation.

Let $f$ be a Horn expression, then using the monotone theory representation (Equation (2)) we know that

$$f = \wedge_{b \in B_H} \mathcal{M}_b(f). \tag{3}$$

Recall that $B_H = \{u \in \{0,1\}^n \mid \text{weight}(u) \geq n - 1\}$, and denote by $b^{(i)}$, $1 \leq i \leq n$, the assignment with $x_i$ set to zero and all other bits set to 1, and by $b^{(0)}$ the assignment $1^n$. In our example $b^{(0)} = 1111$, and $b^{(1)} = 0111$.

Let $D_i$ be the set of clauses that are falsified by $b^{(i)}$, and let $\mathcal{G}_i$ denote the language of all CNF expressions with clauses from $D_i$. In our example, with four variables $a, b, c, d$, clauses in $D_1$ may have $b, c, d$ as negative literals and $a$ as a positive literal. That is, $(\overline{a} \vee \overline{b}) \notin D_1$, but $(a \vee \overline{b}) \in D_1$ and $(\overline{b} \vee \overline{c}) \in D_1$.

Theorem 4 implies that $\mathcal{M}_{b^{(i)}}(f)$ is equal to the least upper bound of $f$ in $\mathcal{G}_i$. Namely, the intersection of all clauses in $D_i$ which are implied by $f$. Define $PI(f, i)$ to be the set of prime implicates of $f$ with respect to $b^{(i)}$. Formally:

$$PI(f, i) = \{d \in D_i | f \models d \text{ and } \forall d' \subseteq d, f \not\models d'\}.$$

Using this notation we get:

$$\mathcal{M}_{b^{(i)}}(f) = \bigwedge_{d \in PI(f,i)} d. \tag{4}$$

Going back to the example $W$, we have:

$$
\begin{aligned}
PI(W, 0) &= (\overline{b} \vee \overline{d})(\overline{c} \vee \overline{d})(\overline{a} \vee \overline{d}) \\
PI(W, 1) &= (\overline{b} \vee \overline{d})(\overline{c} \vee \overline{d}) \\
PI(W, 2) &= (a \rightarrow b)(c \rightarrow b)(\overline{c} \vee \overline{d})(\overline{a} \vee \overline{d}) \\
PI(W, 3) &= (\overline{b} \vee \overline{d})(\overline{a} \vee \overline{d}) \\
PI(W, 4) &= \text{true.}
\end{aligned}
$$

Note that the partition of the prime implicates of $f$ is not disjoint. In particular, the anti-monotone prime implicates (except for $\overline{x_1} \vee \overline{x_2} \vee \ldots \vee \overline{x_n}$ if it is a prime implicate) appear in several $PI(f,i)$ sets. Equation (3) tells us that we can decompose the function into $n + 1$, $b^{(i)}$-monotone functions. Equation (4) tells us how to decompose the clauses of the function, and the monotone theory tells us how to decompose the characteristic models. These observations lead to the following theorem:

**Theorem 10** *The problem CCM, when the input is given as the set of* all *Horn prime implicates, is polynomially equivalent to HTR.*

**Proof:** First observe that the reduction in Theorem 8 uses an anti-monotone function, which has a unique Horn representation. Namely the smallest and the largest representations are the same in this case. This implies that the problem remains as hard as HTR in this special case.

For the other direction, we first partition the input into the sets $PI(f,i)$, and then use a procedure for HTR in order to translate each set to a DNF representation. Then using Claim 1 we translate the DNF expression to the set of minimal assignments. The crucial point is that we have DNF representations for the functions $\mathcal{M}_{b^{(i)}}(f)$ rather than for $f$. This implies that each term in these DNF representations is represented as an element in $char(f)$ and therefore the reduction is polynomial. (We may get some of the elements in $char(f)$ more than once, but at most $n$ times, which is still polynomial.) ∎

In our example, we get the following DNF expressions and their translation into assignments:

$$
\begin{aligned}
PI(W,0) &= \overline{a}\,\overline{b}\,\overline{c} \vee \overline{d} \Rightarrow & \Gamma_0 &= 0001, 1110 \\
PI(W,1) &= \overline{b}\,\overline{c} \vee \overline{d} \Rightarrow & \Gamma_1 &= 0001, 0110 \\
PI(W,2) &= b\overline{d} \vee \overline{ac} \Rightarrow & \Gamma_2 &= 1110, 0001 \\
PI(W,3) &= \overline{a}\,\overline{b} \vee \overline{d} \Rightarrow & \Gamma_3 &= 0001, 1100 \\
PI(W,4) &= \text{true} \Rightarrow & \Gamma_4 &= 1110
\end{aligned}
$$

Similarly we get for SID:

**Theorem 11** *The problem SID, when the output required is* all *Horn prime implicates, is polynomially equivalent to HTR.*

**Proof:** The proof is similar to the proof of the previous theorem. The hardness follows from Theorem 9.

For the other direction, assume we get as input a set $\Gamma$, and an algorithm $A$ for HTR. We first partition $\Gamma$ into sets $\Gamma_i$ according to minimality with respect to $b^{(i)}$. (Note that the sets are not disjoint.) Then we use Claim 1 to transform each $\Gamma_i$ into a DNF expression for the function $\mathcal{M}_{b^{(i)}}(f)$. For each such DNF expression we run the procedure $A$ to compute its CNF representation. By Equation (3), the intersection, with respect to $i$, of these CNF expressions is the Horn expression we need. ∎

In the example, we simply start with the sets $\Gamma_i$ and use the same equations as above going in the other direction. From the above two theorems we get the following corollary.

**Corollary 1** *The problems CCM and SID, when the Horn expression is represented as the set of* all *Horn prime implicates, are polynomially equivalent, and are polynomially equivalent to HTR.*

The equivalence of CCM and SID, in this special case, has been observed before in the database domain (Heikki Mannila, private communication). In fact this led us to the results of this section. As mentioned above a similar result for relational databases is reported by Eiter and Gottlob (1991) where the restriction is called the MAK form for functional dependencies.

**Lifting the Restriction:** The polynomial equivalence to the problem HTR, implies the existence of sub-exponential $n^{O(\log n)}$ algorithm for these problems which may have some practical implications. However, as the following example shows one cannot apply it to solve the general case of the problem SID. Aizenstein and Pitt (1995) present some functions with interesting properties. These functions can be manipulated to create examples with the following properties: (1) $f$ has a short Horn expression, (2) $|char(f)|$ is small, (3) the number of Horn "prime implicates" is exponential. In particular

$$f = (\overline{x_1} \vee \overline{x_2} \vee \ldots \vee \overline{x_m}) \wedge (x1 \vee \overline{y_1}) \wedge (x2 \vee \overline{y_2}) \wedge \ldots \wedge (x_m \vee \overline{y_m})$$

has these properties. The set of prime implicates include all the disjunctions $(b_1 \vee b_2 \vee \ldots \vee b_m)$ where $b_i \in \{\overline{x_i}, \overline{y_i}\}$.

We show by case analysis that the set of characteristic models is small. Observe that in order to satisfy $f$, at least one of the $x_i$ variables must be assigned 0, and that if $x_i = 0$ then $y_i$ must also be assigned 0.

Consider first the set $\min_{1^{2m}}(f)$. Notice that if, for some $j$, $x_j = y_j = 0$ and all the other variables are set to 1, then $f$ is satisfied. This contributes exactly $m$ assignments to $\min_{1^{2m}}(f)$. For $m = 3$ and variable ordering $x_1x_2x_3y_1y_2y_3$, this yields the assignments 011011, 101101, 110110.

Consider next $\min_{b(x_i)}(f)$. Namely, the basis element in which $x_i = 0$. To satisfy $f$, if $x_i = 0$ then $y_i$ must be 0, and as before we can set all other variables to 1. If $x_i = 1$ then there must be another variable $x_j$ which is set to 0. In this case $y_j$ must also be 0. Therefore $\min_{b(x_i)}(f) = \min_{1^{2m}}(f)$.

Lastly, consider $\min_{b(y_i)}(f)$. Namely the basis element in which $y_i = 0$. Observe that $f$ is anti-monotone in $y_i$. Namely, given any satisfying assignment with $y_i = 1$, by flipping $y_i$ to 0 we get another satisfying assignment, which is smaller than the original according to $\leq_{b(y_i)}$. Therefore, we may assume that $y_i = 0$. If $x_i = 0$ then we can set all other variables to 1. If $x_i = 1$ then there must be another variable $x_j$ which is set to 0, and therefore also $y_j = 0$. This assignment is 2 bits away from $b^{(y_i)}$ and it is minimal. We get $m$ assignments in this case too. In our example with $m = 3$, and say $i = 2$, we get the assignments 101101, 011001, and 110100.

Altogether we get $m$ assignments from the first two groups and $m(m-1)$ new assignments from the last and therefore $|char(f)| = m^2$. This means that arbitrary enumeration of the prime implicates, for a given set of models $\Gamma$, is not sufficient for solving SID.

**A Generalization:** While we concentrate in this paper on Horn expressions, we note that the same arguments and proofs hold in the more general case of $k$-quasi Horn expressions. These are expressions in CNF form where in every clause there are at most $k$ positive literals (so that Horn expressions are 1-quasi Horn expressions). The set $B_{H_k} = \{u \in \{0,1\}^n \mid \text{weight}(u) \geq n - k\}$ is a basis for $k$-quasi Horn expressions, and $\Gamma_f^{B_{H_k}}$ can serve as the set of characteristic models for $f$ (Khardon & Roth, 1994). The generalized versions of CCM and SID, when restricted to hold all prime implicates are still equivalent to HTR.

## 4.3 Enumerating Prime Implicants

As mentioned above, given a DNF representation for $f$ we can easily compute the set of characteristic models. One might therefore try to solve CCM by first translating the Horn expression into a DNF expression and then computing the characteristic models from this set. Another possible relaxation is to first compute all the prime implicants of the function and then to extract a DNF representation from it. We consider this problem here. Namely, we consider the problem of enumerating *all* the prime implicants of a Horn expression, and its application for the solution of CCM.

While we have not found a general reduction from this problem to HTR, a simple adaption of the algorithm for HTR (Fredman & Khachiyan, 1994) yields an incremental $n^{O(\log^2 n)}$ algorithm for this problem. However, as we discuss below, enumeration of prime implicants of a Horn expression is not sufficient for solving CCM. The problem in such an application is an exponential gap in the sizes of these representations.

For completeness we sketch the main ideas of the enumeration algorithm here. Let $H$ be a Horn expression, and let $D$ be the DNF expression composed of the prime implicants enumerated so far. The algorithm finds an assignment $x$ which satisfies $H$ and does not satisfy $D$. Using $x$ it is easy to find a new prime implicant of $H$. The algorithm to find $x$ uses the following combinatorial fact (Fredman & Khachiyan, 1994): either there is a variable $x_i$ that appears in $H \wedge \overline{D}$ with high frequency, or the expression $H \wedge \overline{D}$ has "a lot" of satisfying assignments. In the first case, one can recursively solve two sub-problems arrived at by substituting $x_i = 0$, and $x_i = 1$ in the expressions $H$ and $D$. In the second case it is easy to find an assignment $x$ (e.g. by sampling). The solution of the recursion yields the stated time bound. For complete details we refer the reader to the article by Fredman and Khachiyan (1994). While the analysis there is specialized for monotone functions it is easy to extend (the first part of) it for Horn expressions[4].

**Lifting the Restriction:** Denote by $\#PIs(f)$ the number of prime implicants of $f$. While the representations (1) Prime Implicants (PIs), (2) DNF representation, and (3) Characteristic models, satisfy the inequalities $\#PIs(f) \geq |DNF(f)| \geq |char(f)|/n$, each of the inequalities may allow for an exponential gap. The function

$$f_1 \;=\; (\overline{x_1} \vee \overline{x_2} \ldots \vee \overline{x_{\sqrt{n}-1}} \vee x_{\sqrt{n}}) \wedge \ldots \wedge (\overline{x_{n-\sqrt{n}+1}} \vee \overline{x_{n-\sqrt{n}+2}} \vee \ldots \vee \overline{x_{n-1}} \vee x_n)$$

(Khardon & Roth, 1994) shows a gap between (2) and (3). The function

$$f_2 = x_1 x_2 \ldots x_m \vee \overline{x_1}\,\overline{y_1} \vee \overline{x_2}\,\overline{y_2} \vee \ldots \vee \overline{x_m}\,\overline{y_m}$$

(Aizenstein & Pitt, 1995) shows a gap between (1) and (2). (To observe that, notice the similarity between $f_2$ and the dual of the function from the previous sub-section.) Both functions are Horn (for $f_2$ by multiplying out we see that every clause for $f$ is Horn,

---

4. One caveat that we have to tackle is enumerating prime implicants after $D$ is already equivalent to $H$. This can be done using "consensus" operations, which can generate all the prime implicants (Aizenstein & Pitt, 1995) although its Horn expression is large) and both have a small set of characteristic models. These examples show that enumeration of prime implicants may be an inefficient way for producing the characteristic models for some functions.

## 5. A Related Problem

In this section we show that a related problem, which is a minor variant of CCM and SID, is co-NP-Complete. Recall the definition of EOC:

**EOC:** Entailment of Closure
Input: a Horn CNF $H$, a set $G$ of assignments.
Output: Yes if and only if $H \models closure(G)$.

The important difference between CMI and EOC is that the set $G$ is not required to include only satisfying assignments of $H$. This enables the following reduction for EOC, while the complexity of CMI is still open. A similar result in the database domain has been obtained by Gottlob and Libkin (1990).

**Theorem 12** *The decision problem EOC is co-NP-Complete.*

**Proof:** The problem is trivially in co-NP (guess an assignment $x$ and say "No" if $x \in H \setminus closure(G)$).

To show its hardness we reduce co-Monotone 3-SAT to EOC. Monotone 3-SAT (Garey & Johnson, 1979) is the problem of satisfiability of CNF formulas in which in every clause (has 3 literals and) either all the literals are positive (we call these clauses monotone) or all the literals are negated (we call such clauses anti-monotone). Let $f = M \wedge A$ an instance of Monotone 3-SAT where $M$ denotes a conjunction of monotone clauses and $A$ is a conjunction of anti-monotone clauses. We translate it to the instance of EOC: $H = A$ and $\Gamma = \cup_{b \in B_H} min_b(\overline{M})$. First we claim that the reduction is polynomial. Note that since $M$ is a monotone CNF, $\overline{M}$ is a DNF formula in which all the variables are negated, and can therefore be written as an anti-monotone CNF formula. This implies that $\overline{M}$ is Horn, but we have it in a DNF representation. Further computing $\Gamma$ is easy given the DNF representation of $\overline{M}$, and its size is bounded by $(n + 1)$ times the number of clauses in $M$.

We now claim that $f$ is satisfiable if and only if $H \not\models closure(\Gamma)$. Assume first that $f$ is satisfiable, and let $x \in A \wedge M$. This implies that $x \in H$ and $x \notin \overline{M}$. Since $\overline{M}$ is Horn, and the models of Horn functions are closed under intersection (Theorem 1) we get that $x \notin closure(\overline{M})$, and since $\Gamma \subseteq \overline{M}$ $x \notin closure(\Gamma)$. Therefore, $H \not\models closure(\Gamma)$.

For the other direction assume $H \not\models closure(\Gamma)$, and let $x$ be an assignment such that $x \in H$ and $x \notin closure(\Gamma)$. We get that $x \in A$, and since by Theorem 1 and Theorem 2 $\overline{M} = closure(\Gamma)$ we have $x \notin \overline{M}$. So, $x \in A \wedge M$ and $f$ is satisfiable. ∎

To exemplify the above reduction consider the function

$$f = (\overline{a} \vee \overline{b} \vee \overline{c})(\overline{b} \vee \overline{c} \vee \overline{d})(a \vee c \vee d)(a \vee b \vee c).$$

This function will be translated into $H = (\overline{a} \vee \overline{b} \vee \overline{c})(\overline{b} \vee \overline{c} \vee \overline{d})$. The function $\overline{M} = \overline{a}\,\overline{c}\,\overline{d} \vee \overline{a}\,\overline{b}\,\overline{c}$. The satisfying assignments of $\overline{M}$ are $0000, 0001, 0100$, and $\Gamma = char(\overline{M}) = \{0001, 0100\}$. Now consider the assignment $x = 1000$ which satisfies $f$. Clearly, $x$ satisfies $H$, and one can check that it is not in the closure of $\Gamma$.

## 6. Conclusions

Horn expressions and characteristic models are two alternative representations for the same information and none of the two dominates the other in the computational services it can support. The same representations occur in database theory where they have a role in the design of relational databases. A natural question is whether we can translate back and forth between these representations so as to enjoy the benefits of both worlds. In this paper we have studied the computational complexity of these problems.

Our main result is that the two translation problems CCM, and SID, are equivalent to each other (under polynomial reductions), and that they are equivalent to the corresponding decision problem CMI. Namely, translating in either direction is equivalent to deciding whether a given set of models is the set of characteristic models for a given Horn expression.

We have also shown a close relation between our problems and the hypergraph transversal problem HTR. This is a translation problem which is related to many applications in computer science and in particular to AI. We have shown that in general CCM, and SID are at least as hard as HTR, and that in a special case CCM, SID, and HTR are equivalent.

We exhibited examples which show that simple algorithms for enumerating prime implicants cannot guarantee efficient solution for CCM, and similarly enumerating prime implicates may not be efficient for SID. Lastly, we discussed the problem EOC, a minor modification of CMI, which is co-NP-Complete. The complexity hierarchy of the problems discussed is depicted in Figure 1.

Some of the results presented in this paper can be obtained from previous results in database theory, using the equivalence between Armstrong relations and characteristic models reported in a companion paper (Khardon et al., 1995). However, our proofs and exposition make these results much more accessible.

The exact complexity of CMI, and that of HTR are left as open problems. While HTR has a sub-exponential algorithm, the problems CMI might still be co-NP-Hard.

## Acknowledgements

I am grateful to Thomas Eiter, Heikki Mannila, and Dan Roth for their comments which lead to some of the results in this paper. I wish to thank the anonymous referees whose comments helped improve the presentation, and Dimitris Kehagias for his help in proofreading the paper. The research for this paper was supported by Center for Intelligent Control Systems under ARO contract DAAL03-92-G-0115.